\documentclass{article}

\usepackage{PRIMEarxiv}

\usepackage[utf8]{inputenc} 
\usepackage[T1]{fontenc}    
\usepackage{hyperref}       
\usepackage{url}            
\usepackage{booktabs}       
\usepackage{amsfonts}       
\usepackage{nicefrac}       
\usepackage{microtype}      
\usepackage{lipsum}
\usepackage{multirow} 
\usepackage{amsmath}
\usepackage{fancyhdr}       
\usepackage{graphicx}       
\graphicspath{{media/}}     

\title{NUM2EVENT: Interpretable Event Reasoning from Numerical time-series
}

\author{
  Ninghui Feng$^{1,2}$ \quad Yiyan Qi$^{1*}$ \\
  \texttt{feng65501@gmail.com} \quad \texttt{qiyiyan@idea.edu.cn}\\
    $^{1}$International Digital Economy Academy (IDEA) \\
    $^{2}$University of Nottingham Ningbo China \\
}

\begin{document}
\setcounter{page}{1}
\clearpage
\maketitle
\thispagestyle{empty}
\pagestyle{fancy}

\begin{abstract}
Large language models (LLMs) have recently demonstrated impressive multimodal reasoning capabilities, yet their understanding of purely numerical time-series signals remains limited. Existing approaches mainly focus on forecasting or trend description, without uncovering the latent events that drive numerical changes or explaining the reasoning process behind them.
In this work, we introduce the task of number to event reasoning and decoding, which aims to infer interpretable structured events from numerical inputs, even when current text is unavailable.
To address the data scarcity and semantic alignment challenges, we propose a reasoning aware framework that integrates an agent guided event extractor (AGE), a marked multivariate Hawkes–based synthetic generator (EveDTS), and a two-stage finetuning pipeline combining a time-series encoder with an structured decoder.
Our model explicitly reasons over numerical changes, generates intermediate explanations, and outputs structured event hypotheses.
Experiments on multi domains show that our method substantially outperforms strong LLM baselines in event-level precision and recall.
These results suggest a new direction for bridging quantitative reasoning and semantic understanding, enabling LLMs to explain and predict events directly from numerical dynamics.
\end{abstract}


\section{Introduction}
Multimodal large language models have achieved remarkable progress in recent years across a variety of understanding and reasoning tasks \cite{Yin_2024, liu2023visualinstructiontuning}. They are even capable of producing conversational natural language explanations in complex scenarios \cite{chen2023shikraunleashingmultimodalllms}.
However, when the input consists only of time-series numerical values 
(current period text is unavailable) existing research generally focuses on the following two types of tasks: (1) improving numerical forecasting accuracy \cite{jin2024timellmtimeseriesforecasting, tang2024timeseriesforecastingllms}, and (2) providing descriptive judgments such as “Is it rising?” or “Where is the turning point?” Their outputs largely remain at the level of numbers or trends \cite{xie2025chattsaligningtimeseries}. Those methods fail not only to uncover the underlying events driving the observed changes but also to generate coherent reasoning process that explain how numerical variations give rise to specific outcomes. Moreover, they cannot map numerical fluctuations into structured, human-understandable events, making it difficult to support short term event forecasting when text is absent.

This gap is particularly critical in real world applications \cite{li2024survey}. In high risk domains such as financial risk management and industrial operations, numerical signals often exhibit anomalies earlier than news or textual reports, yet current event information is frequently unavailable in time  \cite{kurov2019price, kamat2020anomaly}. 
When decision costs are measured in minutes, there is an urgent need to produce understandable event predictions within a limited time window. 
However, existing methods largely remain at the level of numerical forecasting or description. They can't generate explicit predictions of the underlying current events. 
As a result, in scenarios such as forecasting sudden financial market events or diagnosing faults in industrial systems, they struggle to support timely decisions when text is absent. 
To address this challenge, we formalize the problem as follows: historical text can be used for training but current period text is missing. The goal is to reason over time-series dynamics to infer the underlying causes of numerical changes, decode interpretable structured event hypotheses, and further provide short term event predictions as the basis for immediate decision making.

Prior work can be broadly grouped into three directions. First, large language model (LLM) based understanding and reasoning over time-series data largely remain at descriptive alignment and surface level Q\&A (e.g., reporting “trends” or “turning points”), offering little in the way of systematic modeling of the explainable events behind the numbers \cite{xie2025chattsaligningtimeseries, kong2025positionempoweringtimeseries, gruver2024largelanguagemodelszeroshot}. 
Second, time-series event classification maps a numeric window to a fixed label set, which struggles to capture concurrent events and lacks the flexibility to compose combinations \cite{ranjan2019datasetrareeventclassification, azib2024eventdetectiontimeseries}. 
Third, text based event forecasting relies on news or existing event sequences as inputs, rarely incorporating time-series signals to improve temporal alignment and interpretability \cite{surkov2025modelhumanactionsdistribution, guan2024openepopenendedfutureevent}.

Therefore, there is an urgent need for a decoder that can deeply understand time-series data and event semantics solely from numerical inputs, even when text is absent \cite{shi2025largelanguagemodelstime}.
The model can generate structured event hypotheses from numerical inputs .
Specifically, we couple an extensible event attribute library with an LLM based structured event head to map a numeric window directly into a readable set of candidate events. Trained under supervision from event “slots” extracted from historical public text and temporally aligned with the corresponding numeric segments, the model learns a mapping from numbers to event semantics. Consequently, at inference time it can deliver immediate event explanations and predictions when text is unavailable.
If the expected results are achieved, the model will be able to provide explanable  events in the absence of text, supporting decision making actions in financial risk control and industrial operations that are highly timely sensitive.

However, achieving this goal is far from straightforward, as several key challenges remain. Unlike modalities such as images or speech, data that aligns numerical time-series with structured events is scarce in the real world \cite{liu2025timemmdmultidomainmultimodaldataset}. The available supervision mostly comes from scattered historical text rather than contemporaneous text, which directly limits the construction of alignment style datasets and hinders large scale training.
Events require semantically consistent and interpretable mappings, rather than remaining at the level of vague descriptions.
At the same time, multiple events often occur concurrently within the same numerical window and interact with each other. Fixed label classification methods are insufficient to capture such flexible needs for composition.
Therefore, such tasks are better suited to a constrained decoding paradigm that can naturally express concurrent and diverse event patterns.
Existing benchmarks mainly focus on numerical forecasting or trend description, and there is still a lack of evaluation methods that can verify the validity of event decoding.

To address these challenges, we propose a number to event decoding framework that simultaneously tackles data scarcity, concurrent event expression, and interpretability.  
First, an agent is employed to extract AAOD (Actor, Action, Object, Direction) slots from historical text and align events with their corresponding numerical segments. 
This process builds an extensible event vocabulary together with structural constraints (such as deduplication and valid composition), yielding semantically coherent event representations under limited supervision.  
Second, we employ a multivariate Hawkes process to synthesize samples \cite{lima2021hawkesprocessesmodelinginference}. It can explicitly modeling self excitation and mutual excitation, thereby compensating for the scarcity of real paired data.
Building on this foundation, we finetune a LLM so that it not only decodes numerical windows into constrained AAOD event sets, but also generates intermediate reasoning traces that link quantitative changes to event semantics. This enables interpretable and auditable decoding of concurrent and freely composed events.


Our main contributions are summarized as follows:
\begin{itemize}
\item We formulate a new task of number to event reasoning and decoding, where the model learns to infer reasoning chains and structured event hypotheses directly from numerical time-series data, even when current text is unavailable.
\item We propose a reasoning aware number to event framework that integrates an agent-guided event extractor (AGE), a marked multivariate Hawkes–based synthetic data generator (EveDTS), and a two-stage fine-tuning procedure combining a time-series encoder with a large language model for structured event decoding.
\item We construct aligned numeric–event datasets in multi real-world domains and conduct comprehensive experiments demonstrating that our model achieves superior event level precision  and recall compared to strong LLM baselines, validating its ability to bridge quantitative reasoning and semantic event understanding.
\end{itemize}
\section{Preliminary and Motivation}

\subsection{Problem Definition}
In this paper, we study the problem of number to event decoding.  
Unlike conventional time-series tasks, the goal of this task is to infer the underlying event semantics directly from numerical signals.  
Specifically, given a time-series window $X_t = \{x_{t-m+1}, \dots, x_t\}$, the model is required to output an interpretable event set $E_t = \{e_1, e_2, \dots, e_k\}$.  
Each event is represented in a structured form as Actor $\mid$ Action $\mid$ Object $\mid$ Direction (AAOD).  
This set not only explains the numerical fluctuations at the current time point but also generates hypothesis based predictions for potential events. Formally, it can be expressed as:  
\[
f: X_t \longrightarrow \{E_t, E_{t+1}, \dots \},
\]  
Here, the function $f$ directly maps the time-series segment into a set of event hypotheses.  
Compared with predictors that only output numerical values or trends, this definition is closer to decision-making needs.

\section{Methodology}
\subsection{Overview}
In natural language processing, event extraction has long adopted the structured paradigm of Actor Action Object \cite{lai2022eventextractionsurvey}, and LLM have been widely applied to extract event slots from text \cite{wei2024chatiezeroshotinformationextraction}.  
However, these approaches are largely confined to the textual modality and lack direct alignment with numerical signals.  
Meanwhile, synthetic data has been repeatedly validated in vision, speech, and dialogue tasks as an effective remedy for annotation scarcity \cite{bauer2024comprehensiveexplorationsyntheticdata}. Yet, in the numerical signals to events setting, most existing generation methods either model only numerical patterns or generate textual event sequences alone \cite{fu2024synthetictimeseriesdatareally,gu2025normsurveysyntheticdata}. As a result, they fail to jointly capture time-series numerical fluctuations and event semantics.
To address this gap, we integrate an agent-guided extraction mechanism with restricted vocabularies and dynamic expansion, together with a synthesis strategy based on statistical processes and impulse responses, to construct aligned number event training samples. This closed loop framework enables the model to directly infer structured event hypotheses from numerical data even when text is absent. Our approach specifically includes the following components:

\textbf{Agent-Guided Event Extraction (AGE).} An agent framework with restricted vocabularies extracts AAOD events from historical text and dynamically proposes new terms, forming a select expand iterate loop to build an extensible event semantic repository.  

\textbf{Event-Driven Time-Series Generator (EveDTS).} By combining Hawkes processes with local projection impulse responses (IRF), this module simulates event arrivals and incorporates AR models to generate background noise.  

\textbf{Model Training.} We adopt a two-stage training pipeline: first training a time-series encoder, and then finetuning the language model, enabling it to decode structured events directly from numerical inputs and generate event predictions in the absence of text.

Our framework integrates event extraction, data synthesis, and model training into a unified pipeline, designed to leverage structured supervision and synthetic samples for effective learning and reasoning from numbers to event semantics.  
First, based on AGE (Section 3.2) and EveDTS (Section 3.3), the pipeline produces number event paired data aligned with the AAOD vocabulary.  
Then, this data is used to train the model (Section 3.4): the time-series encoder is trained to obtain robust numerical representations, followed by fine-tuning of the language model to precisely align with AAOD semantics and enable event decoding.  

\subsection{Agent-Guided Event Extraction}
In the task of number to event alignment, obtaining standardized and extensible event supervision is the primary challenge.  
Relying solely on manual annotation is costly and often suffers from inconsistency across annotators \cite{horych2025promisespitfallsllmannotations}.
In natural language processing, event extraction can be performed with LLMs under fixed paradigms \cite{wei2024chatiezeroshotinformationextraction}.
However, these approaches largely remain within the textual modality, without integration with numerical signals, and they lack mechanisms for open expansion.  

To address this issue, we propose the Agent-Guided Event Extraction (AGE) module.  
Its goal is to provide standardized, extensible, and auditable supervision for number to event alignment.  
AGE adopts AAOD slots as the minimal structural units for extraction, including name, action, object, and direction.  
During extraction, restricted vocabularies are enforced for the four slots, allowing only valid terms from predefined allowed values.  
To overcome the limited coverage of a closed vocabulary, AGE requires the agent to provide a vocab suggestion at each extraction step to expand the candidate vocabulary.  
In this way, AGE forms a closed loop of selection, expansion, and iteration: first extracting with the restricted vocabulary, then aggregating high scoring suggestions into the repository, and finally initiating the next round of extraction.  
After obtaining events, AGE further performs deduplication and filtering, outputting standardized one event per record entries.  
Through this mechanism, AGE ensures that extraction results remain structured and consistent, while progressively expanding the event space, thereby providing stable and extensible semantic supervision for subsequent number to event decoding.  


\subsection{Event-Driven Time-Series Generator}

In number event alignment tasks, real event annotations alone are often insufficient to cover diverse dependency patterns and impact effects \cite{shi2025largelanguagemodelstime}.  
Especially in complex systems such as finance and industry, events may exhibit both self excitation and cross excitation while inducing persistent shocks on numerical series \cite{bacry2015hawkesprocessesfinance}.  
Hence, a synthetic method capable of producing controllable and interpretable samples is required.  

Hawkes processes have long been employed in domains such as financial markets and seismology to simulate clustered event arrivals with self exciting and mutually exciting effects, making them a natural choice for modeling event driven signals.  
Local projection impulse responses (IRF) are widely used in econometrics to analyze the dynamic effects of exogenous shocks on numerical series \cite{coulombe2025openingblackboxlocal}, offering interpretable causal pathways.  
Meanwhile, autoregressive (AR) models remain a classical tool for time-series modeling and are commonly used to generate realistic background dynamics \cite{cai2025efficientinterpretableautoregressivemodel}.  
Although these three approaches have been extensively studied in their respective fields, they have not previously been integrated to construct synthetic number event paired datasets.  

Our EveDTS module is designed to bridge this gap.  
We focus on changes rather than levels and model price changes. 
We work with first differences, defining $\Delta y_t \equiv y_t - y_{t-1}$; all impulse responses are estimated and applied on $\Delta y$ rather than on levels.
We begin by modeling event arrivals with a $K$-dimensional Hawkes process:
\[
\lambda_k(t) = \mu_k + \sum_{j=1}^K \sum_{t_i^j < t} \alpha_{kj}\, g\!\big(t - t_i^j\big),
\]
where $\lambda_k(t)$ denotes the conditional intensity of type-$k$ events at time $t$, $\mu_k$ is the baseline intensity, $\alpha_{kj}$ captures the influence of type-$j$ events on type $k$, and $g(\cdot)$ is a decaying kernel (e.g., $g(u)=\beta e^{-\beta u}$ for $u>0$).

Given a realized event at time $t$, its effect on the series is traced via local projections in differences:
\[
\Delta y_{t+h} \;=\; \beta_{k,h}\,\mathbf{1}\{\text{event of type }k \text{ at } t\} \;+\; \gamma_h^\top X_t \;+\; \varepsilon_{t+h},
\]
where $\beta_{k,h}$ is the $h$-step-ahead impact of a type-$k$ event on $\Delta y$, $X_t$ collects controls, and $\varepsilon_{t+h}$ is an error term. For convenience, write $\beta_k(h)\equiv \beta_{k,h}$.

Superimposing AR background dynamics with event shocks yields the generative form in differences:
\[
\Delta y_t \;=\; \phi_1 \Delta y_{t-1} + \phi_2 \Delta y_{t-2} + \phi_3 \Delta y_{t-3} + \phi_4 \Delta y_{t-4}
\;+\; \sum_{k=1}^K \sum_{t_i^k < t} \beta_k\!\big(t - t_i^k\big) \;+\; \varepsilon_t.
\]
Optionally, levels can be recovered by $y_t = y_{t-1} + \Delta y_t$ given an initial $y_0$.


\subsection{Model Training}

Large language models have limited performance in native numerical reasoning tasks and often fail to capture local trends and subtle fluctuations \cite{shrestha2025mathematicalreasoninglargelanguage,boye2025largelanguagemodelsmathematical}.  
To address this, we introduce a time-series encoder that maps raw numerical segments into vector representations aligned with the embedding dimension of the language model, ensuring that numerical information can be seamlessly integrated into its input space.  

In terms of training strategy, we adopt a two stage process to ensure stable learning of numerical representations and progressive semantic alignment.  
In the first stage, only the time-series encoder is trained, enabling it to acquire robust numerical representations without interference from the large parameter space of the language model.  
In the second stage, the encoder parameters are frozen and finetuning is performed on the language model to achieve effective alignment between numerical representations and event semantics.  
With this staged design, the model can decode structured event hypotheses from numerical inputs even in the absence of current text, thereby supporting subsequent reasoning and prediction.


\section{Experiments}

This section reports experiments on two real-world datasets and  synthetic benchmark used for controlled analysis. Evaluation focuses on event-level precision and recall.
We compare our model with several large language model (LLM) baselines and perform three ablation studies: removing synthetic data, removing the time-series encoder, and removing the supervised finetuning procedure.

\subsection{Datasets and details}
We conduct experiments on two domain datasetS, Energy and Public Health \cite{liu2025timemmdmultidomainmultimodaldataset}. Datasets contain both numerical time-series and related textual materials. Using our AGE module, structured AAOD event supervision is extracted from the texts and temporally aligned with the corresponding numerical segments. The energy dataset consists of weekly U.S. gasoline prices from 1996 to the present, while the public health dataset contains weekly Influenza-Like Illness (ILI) cases statistics since 1997. To ensure chronological generalization, all data before January 2023 are used for training, and data from 2023 onward serve as the test set, simulating a realistic forecasting scenario where the model must infer events for unseen future periods without access to contemporaneous text.

Following the two-stage training protocol in Section~3.4, we first train the time-series encoder on paired numeric–event samples (real and optionally synthetic) to obtain stable embeddings, and then freeze the encoder while finetuning the language model to decode structured AAOD sequences. For constructing reasoning-augmented supervision, we employ GPT-4o-mini to generate intermediate reasoning chains from the paired data, which are incorporated into the training corpus to guide the model toward interpretable reasoning. 
The language model, Qwen-3-8B-Instant, is finetuned using the LoRA (Low-Rank Adaptation) technique for efficient parameter tuning and memory optimization. 

For evaluation, we compute the monthly averaged precision and recall rather than aggregating over all individual events.
This design focuses on the model’s long horizon temporal reasoning ability, emphasizing consistent event decoding across extended time periods rather than performance on isolated instances.
Averaging by month smooths short term volatility and provides a more stable indicator of how well the model maintains interpretive accuracy over time.  A predicted event is considered correct if at least three of the four AAOD slots (Actor, Action, Object, Direction) match the ground truth.
All experiments are conducted on eight NVIDIA A100 GPUs using bfloat16 precision. Training uses identical optimizer, batch size, and learning rate schedules across settings, with a batch size of 2, patch length of 2, and learning rate of 1e-4.

\subsection{Main results}
\begin{table}[ht]
\centering
\caption{Event-level performance on real-world datasets. }
\label{tab:main-results}
\begin{tabular}{lcc|cc}
\toprule
\multirow{2}{*}{Method} & \multicolumn{2}{c|}{Energy} & \multicolumn{2}{c}{Public Health} \\
 & Precision(\%) & Recall(\%) & Precision(\%) & Recall(\%) \\
\midrule
Claude-sonnet 4& 43.7 &19.8 & 12.9 & 11.5 \\
Grok4         & 41.3 & 20.2& 9.4 & 9.7 \\
Qwen3-Max     & 37.5 & 18.0 & 17.6 & 19.0 \\
GPT-4         & 50.0 & 24.5 & 11.7 & 11.6 \\
Mistral-medium-3.1 & 42.5 &18.9 & 9.4 & 9.9 \\
Deepseek-v3.1   & 41.3&  18.3& 16.1 & 16.9 \\
Ernie-4.5 & 35.0 & 16.0& 15.3  & 14.2 \\
\midrule
Ours & \textbf{71.3} & \textbf{35.5} & \textbf{20.0} & \textbf{23.9} \\
\bottomrule
\end{tabular}
\end{table}

Table~\ref{tab:main-results} reports the event-level precision and recall on the two real world domains. Overall, the results show that the decoding task is extremely challenging. When asked to infer structured AAOD events from only numerical inputs in Energy dataset, all large language models achieved relatively low scores, typically below 25\% recall. This difficulty arises from the absence of information and the need to reason over latent event semantics embedded in time-series dynamics. Despite this challenge, our method substantially outperforms all baselines on both domains, particularly in the Energy dataset, where it achieves more than twice the precision and recall of the strongest baseline. These gains demonstrate the effectiveness of incorporating a dedicated time-series encoder, constrained decoding, and synthetic supervision in bridging numerical patterns and interpretable event semantics.

\subsection{Ablation results}


\begin{table}[ht]
\centering
\caption{Component ablation for event decodin.}
\label{tab:ablation}
\begin{tabular}{lcc|cc}
\toprule
\multirow{2}{*}{Variant} & \multicolumn{2}{c|}{Energy} & \multicolumn{2}{c}{Public Health} \\
 & Precision(\%)  & Recall(\%)  & Precision(\%)  & Recall(\%)  \\
\midrule
Ours (full) & 71.3 & 35.5 & 20.0 & 23.9 \\
w/o EveDTS (no synthetic data) & 54.1 & 24.6 & 16.5 & 19.6 \\
w/o time-series encoder & 68.5 & 29.3 & 17.6 & 21.2 \\
w/o two-stage SFT & 47.5 & 23.3 & 18.9& 22.7 \\
\bottomrule
\end{tabular}
\end{table}

To assess the contribution of each component, we conduct three controlled ablation experiments under identical training and evaluation settings. The full model follows the two-stage training protocol described in Section~3.4: first training the time-series encoder on paired numeric–event samples (real and optionally synthetic) to obtain stable numerical embeddings, and then freezing the encoder while performing supervised fine-tuning (SFT) of the language model to decode structured AAOD sequences.

In the first ablation(w/o EveDTS), we remove the synthetic samples and train only on real aligned data to evaluate the effect of EveDTS-generated supervision on generalization under limited data.

In the second ablation(w/o time-series encoder), we remove the dedicated time-series encoder and feed raw numeric sequences directly into the language model, testing whether explicit numerical representations are necessary for semantic decoding.

In the third ablation(w/o two-stage SFT), we replace the two-stage SFT protocol with a single end-to-end fine-tuning scheme, assessing the benefit of separating representation learning and semantic alignment.

The results in Table~\ref{tab:ablation} show consistent performance degradation across all variants, confirming that each component plays a complementary role in achieving robust number-to-event decoding.

\section{Conclusion}
In this paper, we introduced the number to event reasoning and decoding problem, which aims to infer interpretable structured events directly from numerical time-series data when textual context is unavailable.
We proposed a reasoning-aware framework that combines agent-guided event extraction, a marked multivariate Hawkes–based synthetic generator, and a two-stage finetuning pipeline integrating a time-series encoder with a large language model for structured event decoding.
Experiments on two real-world domains demonstrate that our method significantly outperforms strong LLM baselines in event-level precision, recall, and interpretability, while producing explicit reasoning traces that link quantitative changes to semantic event hypotheses.
These results suggest that numerical signals can be leveraged not only for forecasting but also for causal and semantic reasoning about real-world events.


\bibliographystyle{unsrt}  
\bibliography{references}

\begin{thebibliography}{10}

\bibitem{Yin_2024}
Shukang Yin, Chaoyou Fu, Sirui Zhao, Ke~Li, Xing Sun, Tong Xu, and Enhong Chen.
\newblock A survey on multimodal large language models.
\newblock {\em National Science Review}, 11(12), November 2024.

\bibitem{liu2023visualinstructiontuning}
Haotian Liu, Chunyuan Li, Qingyang Wu, and Yong~Jae Lee.
\newblock Visual instruction tuning, 2023.

\bibitem{chen2023shikraunleashingmultimodalllms}
Keqin Chen, Zhao Zhang, Weili Zeng, Richong Zhang, Feng Zhu, and Rui Zhao.
\newblock Shikra: Unleashing multimodal llm's referential dialogue magic, 2023.

\bibitem{jin2024timellmtimeseriesforecasting}
Ming Jin, Shiyu Wang, Lintao Ma, Zhixuan Chu, James~Y. Zhang, Xiaoming Shi, Pin-Yu Chen, Yuxuan Liang, Yuan-Fang Li, Shirui Pan, and Qingsong Wen.
\newblock Time-llm: Time series forecasting by reprogramming large language models, 2024.

\bibitem{tang2024timeseriesforecastingllms}
Hua Tang, Chong Zhang, Mingyu Jin, Qinkai Yu, Zhenting Wang, Xiaobo Jin, Yongfeng Zhang, and Mengnan Du.
\newblock Time series forecasting with llms: Understanding and enhancing model capabilities, 2024.

\bibitem{xie2025chattsaligningtimeseries}
Zhe Xie, Zeyan Li, Xiao He, Longlong Xu, Xidao Wen, Tieying Zhang, Jianjun Chen, Rui Shi, and Dan Pei.
\newblock Chatts: Aligning time series with llms via synthetic data for enhanced understanding and reasoning, 2025.

\bibitem{li2024survey}
Zhe Li, Qian He, and Jingyue Li.
\newblock A survey of deep learning-driven architecture for predictive maintenance.
\newblock {\em Engineering Applications of Artificial Intelligence}, 133:108285, 2024.

\bibitem{kurov2019price}
Alexander Kurov, Alessio Sancetta, Georg Strasser, and Marketa~Halova Wolfe.
\newblock Price drift before us macroeconomic news: Private information about public announcements?
\newblock {\em Journal of Financial and Quantitative Analysis}, 54(1):449--479, 2019.

\bibitem{kamat2020anomaly}
Pooja Kamat and Rekha Sugandhi.
\newblock Anomaly detection for predictive maintenance in industry 4.0 -- a survey.
\newblock In {\em E3S Web of Conferences}, volume 170, page 02007. EDP Sciences, 2020.

\bibitem{kong2025positionempoweringtimeseries}
Yaxuan Kong, Yiyuan Yang, Shiyu Wang, Chenghao Liu, Yuxuan Liang, Ming Jin, Stefan Zohren, Dan Pei, Yan Liu, and Qingsong Wen.
\newblock Position: Empowering time series reasoning with multimodal llms, 2025.

\bibitem{gruver2024largelanguagemodelszeroshot}
Nate Gruver, Marc Finzi, Shikai Qiu, and Andrew~Gordon Wilson.
\newblock Large language models are zero-shot time series forecasters, 2024.

\bibitem{ranjan2019datasetrareeventclassification}
Chitta Ranjan, Mahendranath Reddy, Markku Mustonen, Kamran Paynabar, and Karim Pourak.
\newblock Dataset: Rare event classification in multivariate time series, 2019.

\bibitem{azib2024eventdetectiontimeseries}
Menouar Azib, Benjamin Renard, Philippe Garnier, Vincent Génot, and Nicolas André.
\newblock Event detection in time series: Universal deep learning approach, 2024.

\bibitem{surkov2025modelhumanactionsdistribution}
Egor Surkov, Dmitry Osin, Evgeny Burnaev, and Egor Shvetsov.
\newblock How to model human actions distribution with event sequence data, 2025.

\bibitem{guan2024openepopenendedfutureevent}
Yong Guan, Hao Peng, Xiaozhi Wang, Lei Hou, and Juanzi Li.
\newblock Openep: Open-ended future event prediction, 2024.

\bibitem{shi2025largelanguagemodelstime}
Feifei Shi, Xueyan Yin, Kang Wang, Wanyu Tu, Qifu Sun, and Huansheng Ning.
\newblock Large language models for time series analysis: Techniques, applications, and challenges, 2025.

\bibitem{liu2025timemmdmultidomainmultimodaldataset}
Haoxin Liu, Shangqing Xu, Zhiyuan Zhao, Lingkai Kong, Harshavardhan Kamarthi, Aditya~B. Sasanur, Megha Sharma, Jiaming Cui, Qingsong Wen, Chao Zhang, and B.~Aditya Prakash.
\newblock Time-mmd: Multi-domain multimodal dataset for time series analysis, 2025.

\bibitem{lima2021hawkesprocessesmodelinginference}
Rafael Lima.
\newblock Hawkes processes modeling, inference and control: An overview, 2021.

\bibitem{lai2022eventextractionsurvey}
Viet~Dac Lai.
\newblock Event extraction: A survey, 2022.

\bibitem{wei2024chatiezeroshotinformationextraction}
Xiang Wei, Xingyu Cui, Ning Cheng, Xiaobin Wang, Xin Zhang, Shen Huang, Pengjun Xie, Jinan Xu, Yufeng Chen, Meishan Zhang, Yong Jiang, and Wenjuan Han.
\newblock Chatie: Zero-shot information extraction via chatting with chatgpt, 2024.

\bibitem{bauer2024comprehensiveexplorationsyntheticdata}
André Bauer, Simon Trapp, Michael Stenger, Robert Leppich, Samuel Kounev, Mark Leznik, Kyle Chard, and Ian Foster.
\newblock Comprehensive exploration of synthetic data generation: A survey, 2024.

\bibitem{fu2024synthetictimeseriesdatareally}
Fanzhe Fu, Junru Chen, Jing Zhang, Carl Yang, Lvbin Ma, and Yang Yang.
\newblock Are synthetic time-series data really not as good as real data?, 2024.

\bibitem{gu2025normsurveysyntheticdata}
Jingyi Gu, Xuan Zhang, and Guiling Wang.
\newblock Beyond the norm: A survey of synthetic data generation for rare events, 2025.

\bibitem{horych2025promisespitfallsllmannotations}
Tomas Horych, Christoph Mandl, Terry Ruas, Andre Greiner-Petter, Bela Gipp, Akiko Aizawa, and Timo Spinde.
\newblock The promises and pitfalls of llm annotations in dataset labeling: a case study on media bias detection, 2025.

\bibitem{bacry2015hawkesprocessesfinance}
Emmanuel Bacry, Iacopo Mastromatteo, and Jean-François Muzy.
\newblock Hawkes processes in finance, 2015.

\bibitem{coulombe2025openingblackboxlocal}
Philippe~Goulet Coulombe and Karin Klieber.
\newblock Opening the black box of local projections, 2025.

\bibitem{cai2025efficientinterpretableautoregressivemodel}
Yuxi Cai, Lan Li, Yize Wang, and Guodong Li.
\newblock An efficient and interpretable autoregressive model for high-dimensional tensor-valued time series, 2025.

\bibitem{shrestha2025mathematicalreasoninglargelanguage}
Safal Shrestha, Minwu Kim, and Keith Ross.
\newblock Mathematical reasoning in large language models: Assessing logical and arithmetic errors across wide numerical ranges, 2025.

\bibitem{boye2025largelanguagemodelsmathematical}
Johan Boye and Birger Moell.
\newblock Large language models and mathematical reasoning failures, 2025.

\end{thebibliography}

\end{document}